\documentclass{article}
\usepackage[T1]{fontenc}
\usepackage[utf8]{inputenc}
\usepackage{graphicx}
\usepackage{color,soul}
\usepackage{url} 
\usepackage{hyperref}
\usepackage{adjustbox}
\usepackage[margin=4cm]{geometry}
\hypersetup{
   colorlinks,
   citecolor=blue,
   filecolor=blue,
   linkcolor=blue,
   urlcolor=blue
}

\begin{document}
\title{A Search for Improved Performance \\ in Regular Expressions}
\author{\small Brendan Cody-Kenny \thanks{Brendan Cody-Kenny, Michael Fenton, and Michael O'Neill are with the Natural Computing Research and Applications Group (NCRA) in University College Dublin, Ireland. Email: \href{mailto:brendan.cody-kenny@ucd.ie}{Brendan.Cody-Kenny@ucd.ie}, \href{mailto:michaelfenton1@gmail.com}{MichaelFenton1@gmail.com}, \href{mailto:m.oneill@ucd.ie}{M.ONeill@ucd.ie}.}
  \and \small Michael Fenton \footnotemark[1]
  \and \small Adrian Ronayne \thanks{Adrian Ronayne, Eoghan Considine and Thomas McGuire, are with the Fidelity Center for Applied Technology (FCAT) in Fidelity Investments, Ireland. Email: \href{mailto:Adrian.Ronayne@fmr.com}{Adrian.Ronayne@fmr.com}, \href{mailto:eoghan.considine@fmr.com}{Eoghan.Considine@fmr.com}, \href{mailto:Thomas.McGuire@fmr.com}{Thomas.McGuire@fmr.com}.}
  \and \small Eoghan Considine \footnotemark[2]
  \and \small  Thomas McGuire \footnotemark[2]
  \and \small  Michael O'Neill \footnotemark[1]
}

\maketitle

\begin{abstract}
\label{abstract}

The primary aim of automated performance improvement is to reduce the running time of programs while maintaining (or improving on) functionality. In this paper, Genetic Programming is used to find performance improvements in regular expressions for an array of target programs, representing the first application of automated software improvement for run-time performance in the Regular Expression language. This particular problem is interesting as there may be many possible alternative regular expressions which perform the same task while exhibiting subtle differences in performance. A benchmark suite of candidate regular expressions is proposed for improvement. We show that the application of Genetic Programming techniques can result in performance improvements in all cases.

As we start evolution from a known good regular expression, diversity is critical in escaping the local optima of the seed expression. In order to understand diversity during evolution we compare an initial population consisting of only seed programs with a population initialised using a combination of a single seed individual with individuals generated using PI Grow and Ramped-half-and-half initialisation mechanisms.

\end{abstract}

\section{Introduction}
\label{sec:intro}

The automatic improvement of existing software has gained considerable ground in recent years \cite{white:2011:evolutionary,langdon:2015:optimizing}. The goal of automated software improvement is to modify program code such that overall performance (as measured by some metric, usually execution time) is improved, while maintaining functional properties (or in some cases improving them also) as measured by a test suite \cite{langdon:2015:optimizing,cody2015locogp}. 

Regular Expressions (regexes) are strings of literal and special characters that are used to match sub-strings in text \cite{thompson1968programming}. 
The use of special characters can result in regexes which are considered ``correct'' but suffer from severe performance issues. 
There may exist multiple different implementations which achieve the same or very similar functionality. 
A regex may be deployed and executed in different environments, with different performance characteristics being exhibited as it is interpreted by different regex engines. Different interpreter implementations of the same language can vary in performance \cite{selakovic2016performance}. This variability motivates an automated approach to tease out performance improvements, particularly where improvements may be unexpected or counter-intuitive. 

While our overall goal is to improve performance, work on multi-version programming is highly relevant as we are seeking functionally equivalent programs which have alternate implementations \cite{feldt1998generating}, but which also happen to execute faster. The problem of performance improvement is thus split into two smaller problems:

\begin{enumerate}
	\item finding alternate implementations with equivalent functionality, and
	\item reducing unnecessary execution from the given target.
\end{enumerate}

In this paper, we evaluate how Genetic Programming (GP) can be used to improve the performance of regexes in terms of minimising execution or ``wall-clock'' time. An automated search process such as GP is ideal for exploring a large range of program variants which exhibit subtle performance differences \cite{koza:2003:GPIV}. In order to examine the difficulty of the space, we have compiled a benchmark suite of target regular expressions for an array of common regex applications, including MAC address search and validation \cite{bartoli2016inference} and email validation \cite{AngularJS}. Using Grammatical Evolution (GE) \cite{o2003grammatical}, a grammar-based form of Genetic Programming (GP) \cite{koza:2003:GPIV}, we explore many variants of these regexes in the search for improvements which yield reduced execution time. 

The topic of performance improvement of regular expressions using a population-based evolutionary search algorithm raises an interesting question: are common regexes available from source code repositories capable of being improved at all? Furthermore, is it possible to generate syntactically distinct solutions while maintaining functionality?

This work describes a pilot study in the domain of automated regular expression improvement, which forms a basis for future work in the area. Furthermore, there is a significant gap in the literature; as far as the authors are aware, this work represents the first application of automated software improvement in the regular expression domain using heuristic algorithms such as GP. 

The rest of this paper is structured as follows. Section \ref{sec:related_work} gives an overview of the background to both regular expressions (Section \ref{subsec:regex_background}) and software improvement (Section \ref{subsec:software_improvement_background}). Our methodology is detailed in Section \ref{sec:method}, including our approach to population seeding with target solutions (Section \ref{subsec:pop_seeding}) and a detailed description of our fitness function in Section \ref{subsec:fitness_function}. Experiments are outlined in Section \ref{sec:experiments}, with our benchmark test suite of target regexes for improvement detailed in Section \ref{subsec:regex_test_suite}. Our results and discussion thereof are presented in Section \ref{sec:results}. Finally, our conclusions are drawn in Section \ref{sec:conclusions}, and we present avenues for future work in Section \ref{subsec:future_work}.

\section{Related Work}
\label{sec:related_work}

Program execution time can be difficult to measure and reason about for a number of reasons \cite{st2015optimization}. 
Reading through program source code can only partially indicate expected performance as many other aspects have an impact on the time it takes for a program to finish.
Firstly, the size and distribution of data passed to a program can sometimes induce an exponential increase in the run-time cost of execution after a program has been deployed \cite{nistor2014suncat}. This is problematic as the input data sizes used to test a program during development may not be representative of the programs use once deployed. There is less control of what data is encountered once a program is deployed.
Secondly, the run-time environment can obscure accurate performance measurements as computing resources are shared and scheduled among other programs. Programs may be subject to optimisation strategies such as just-in-time compilation \cite{st2015optimization}. Where a program must operate across a number of different run-time environments it may be even more difficult to elucidate program performance on differing architectures, libraries and APIs~\cite{selakovic2016performance}. In such widely varying circumstances a single program execution will not give an accurate performance measurement. To gather accurate measures many repeated program executions are needed which yield a range of values. As a result of these issues, performance improvements are frequently made opportunistically or in response to glaring performance bottlenecks \cite{linares2015developers} and as such, are frequently made in an ad-hoc manner without any rigorous analysis of performance \cite{selakovic2016performance}. 

The use of evolutionary techniques on regular expressions has so far largely focused on evolving regexes in terms of their functionality and readability. As such, performance remains an important open issue in writing regexes. We draw on a number of previous works which demonstrate Genetic Improvement of existing programs for performance in other languages such as C++ \cite{langdon2013faster,petke2014using} and Java \cite{cody2015locogp}.

\subsection{Evolving Regular Expressions}
\label{subsec:regex_background}

It can be difficult to write regexes which capture all edge cases \cite{cochran2015program}. ``Program Boosting'' in this context refers to the use of GP to merge many human-written regexes into a single expression which captures all edge cases. From a Software Engineering perspective, this work is very interesting in that many people can work in isolation on a solution, coming up with novel and unique ``building blocks'' which are reworked and merged by evolutionary approaches into a regex which better fulfills the overall requirements. If writing functional code were not hard enough, non-functional requirements (e.g. run-time or memory) introduce complex trade-offs which must also be considered. Further exacerbating the issue is that the environment in which a program must operate can vary widely even between different versions of the same language interpreter \cite{selakovic2016performance}.

Bartoli \emph{et al.} showed that it is possible to evolve working regexes \cite{bartoli2016inference}. Evolving regexes which pass all tests is possible in a reasonable amount of time (minutes) without seeding evolution with known useful partial regexes. For example, the string

\begin{center}
\texttt{eth0 119.63.193.196(5c:0a:5b:63:4a:82):4399}
\end{center}

\noindent contains a MAC address consisting of six pairs of hex characters in the ranges 0-9 and a-f with each pair separated by a colon. A correct match will return the string \verb;5c:0a:5b:63:4a:82; as well as the start and end positions (20,36). Regexes are evolved and tested against a set of similar test strings. A fitness gradient is provided by counting the number of match errors which an evolved regexes returns. A regex which returns either \verb;(5c:0a:5b:63:4a:82); or \verb;5c:0a:5b:63:4a:; is off by 2 characters.

The search algorithm proposed by Bartoli \emph{et al.} applies GP multiple times on test cases derived from the example shown above \cite{bartoli2016inference}. The idea here is to use GP to find regex elements which match some small portion of the desired text. An interesting point is that the desired text is decomposable into sub-problems by considering smaller portions of the desired text, something that is not straightforward for more complex programs. If a regex can match the first 3 characters of a MAC address then that regex can be considered a building block for the overall solution. These partial ``building block'' regexes are composable with the ``or'' operator \verb;|;. The decomposability of the test cases as well as the composability of the solution appears a key enabler in evolving regexes relatively quickly. 

Subsequent to evolving and composing regexes, GP has also proven capable of evolving shorter, and therefore presumably more readable regexes \cite{bartoli2014playing}. This highlights that GP is very useful for reducing a large, complex or general program down into one which is more specialized for a particular case. While regexes are evolved to be functionally correct, shorter, and more readable, run-time performance is not taken into account \cite{bartoli2016inference}. Deterministic approaches to ``optimise'' regexes create more concise versions while maintaining semantic equivalence with rule-based approaches \cite{regexp-optimizer,regexp-assemble}.

If we are to contrast the evolution of regex functionality with evolution of regex performance, finding improvements which show up as a decrease in execution time is not as closely related to the makeup of each part of a regex. For this reason we do not expect that the composability of regexes to be as important for improving run-time as it is for ``growing'' functionality. 

\subsection{Performance Improvement in Software}
\label{subsec:software_improvement_background}
In this paper we use ``performance'' to refer to time elapsed for some execution. Reliably measuring wall-clock time or ``benchmarking'' programs is subject to variability introduced by the execution environment, especially for interpreted languages.
When improving existing software, we are starting with a known ``good''  initial program which is near-optimal or mostly correct.
As the initial program already embodies a mostly acceptable program, it represents a local optima which evolution must escape.
The initial existing program can be seeded into the first generation and/or mutated to initialize the first generation \cite{cody2015locogp} of an evolutionary algorithm.

Previous work in program improvement demonstrated improvements to programs resulting from a small number of modifications to code \cite{cody2015locogp} as well as through the removal of unnecessary code \cite{langdon:2015:optimizing}. One of the most effective ways to reduce the execution time of a program is simply to remove redundant executions \cite{langdon:2015:optimizing}.
   Improved program variants have been found where up to 15\% of the code is modified under mutation \cite{petke2014using}, however the lion's share of execution reduction can often be achieved with fewer modifications to the code \cite{cody2015locogp,langdon:2015:optimizing,petke2014using}. Though software has been shown relatively robust to mutation \cite{schulte2014software}, diversity can be an issue, particularly when mutating small programs in a strongly-typed language such as Java \cite{cody2015locogp}. When starting from an existing program we are essentially seeking alternative implementations of functionality, and for this reason, finding diverse implementations is of interest.

\subsection{Diverse Implementations}
\label{subsec:diversity}
Ideally we would like to enumerate and compare performance of multiple diverse implementations which provide the same functionality. 
The most prominent means for promoting diversity in evolutionary algorithms utilize multi-objective optimisation to drive both fitness and diversity in a population \cite{laumanns:2002:combining,deb:2002:NSGA2}. Other approaches seek to improving diversity by introducing known relevant building blocks from a range of existing programs \cite{petke2014using}.
Diversity has also been strongly encouraged by using multiple GP runs with different configurations over a range different primitives to find entirely different implementations of the same functionality in the context of multi-version programming \cite{feldt1998generating}.  

\section{Method}
\label{sec:method}

We use Grammatical Evolution as implemented\footnote{Source code and experimental configurations used in this paper are available at \href{https://github.com/codykenb/PonyGE2}{https://github.com/codykenb/PonyGE2}, specifically commit \href{https://github.com/codykenb/PonyGE2/tree/2e7fa0184b69cca31c078963e58857c9f563d20e}{2e7fa0184b69cca31c078963e58857c9f563d20e}.}
 in the ``PonyGE2'' evolutionary framework \cite{PonyGE2}.
 In order to make regular expressions amenable to evolution in PonyGE2, a GE-compatible LR parser was constructed for this application \cite{ReverseGE}. This parser quickly converts any target regex string using a given grammar and returns a fully parsed, PonyGE2-compatible tree. 

\subsection{Initialisation \& Population Seeding}
\label{subsec:pop_seeding}

Initialisation typically has a large effect on the overall performance of the evolutionary algorithm \cite{koza:2003:GPIV}.
We compare three different methods of composing the initial population. The first method uses only the seed regex. The other two methods use a single seed along with more traditional initialisation techniques.  

\textbf{Seed Only}

\noindent As the target program is known, we investigate initialising the first generation using only the seed program. While subtree crossover will have no effect on the initial generation (since all individuals will be identical), subtree mutation will inject new genetic material into the population. Diversity will increase in subsequent generations as offspring programs are subjected to further mutation and crossover. As the initial seed program is already highly fit (i.e. it should pass all test cases by default), the search process will first need to navigate its way out of this local optima. 

\textbf{Ramped Half-and-Half + Seed}

\noindent Ramped Half-and-Half (RHH) is a commonly used initialisation method for GP-based techniques \cite{koza1992genetic}. This approach generates a population of individuals across a range of derivation tree depths. The tree depth is ``ramped'' up from zero to produce a range of program sizes. At each depth, derivation trees are created using ``full'' (every branch in the tree is forced to the given depth) and ``grow'' (no branch in the tree is forced to the depth and the tree is allowed to grow at random up to that given depth) derivation methods. While this technique is widely used, there have been concerns over its appropriateness in certain applications \cite{luke2001survey, harper2010ge, fagan:2016:PI, Nicolau:2016:Repitition}. A single seed individual program is then added to the initialised population, resulting in the population instantly gaining a highly fit local optimum. In comparison to ``seed only'', RHH introduces an increased amount of `noise' in the population which may make escaping this local optimum more difficult. 

\textbf{PI Grow + Seed}

\noindent Position Independent Grow (PI Grow) has been proposed as a viable alternative to RHH initialisation \cite{fagan:2016:PI}. Whereas RHH generates pairs of trees at a range of depths, PI Grow eschews the combination of full and grow derivations and generates individuals at a range of depths where at least one branch of the derivation tree is forced to the given depth. Furthermore, to combat the leftmost derivation tendencies of pre-fix or in-fix grammar-based mapping systems, PI Grow derives trees in a position independent manner by randomising the order of derivation of non-terminals \cite{fagan:2016:PI}. This has the effect of reducing inherent biases which are intrinsic to grammar-based systems. As with RHH above, a single instance of the ``local optima'' seed individual program is included into the initialised population.

\subsection{Fitness Function}
\label{subsec:fitness_function}

Our fitness function is a sum of functionality errors and execution time in milliseconds.

\textbf{Execution Time} is a sum measure of the time a regex takes on multiple input values (test cases). We use the timeit Python library which makes certain provisions for accurately timing execution such as temporarily disabling the Python garbage collector \cite{timeit}. We take the best of 3 repititions for each test case. Occasionally, evolved regexes will exhibit exponential execution behaviour due to catastrophic backtracking, needlessly delaying the run. To mitigate this issue we impose a one second timeout for any given regex to complete all tests. As a result, execution time can not contribute more than 1 to the fitness. One second is many times longer than the time it takes each regex to search through all test cases.

\textbf{Functionality Error} is summed across many positive and negative test cases. Errors within a positive matching test case are a sum of the the number of incorrect characters matched as well as the number of missing characters per match. A missing character is weighted the same as an incorrectly matched character. Where a number of matches are expected, for example where multiple MAC addresses exist within a string, the number of incorrect matches are added to the functionality. The minimum measure of functionality error is one, meaning any regex which contains a detectable error can not receive a fitness value better than a regex which contains no error. 

\section{Experiments}
\label{sec:experiments}

All experiments use ``PonyGE2'' \cite{PonyGE2} with experimental parameters as summarised in Table~\ref{tab:gpconfig}. We conduct a single run to 1000 generations for each problem to find some interesting ``best found'' examples of our regexes. For subsequent results in comparing initialisation methods, we found 100 generations to be enough to find improvements. A bootstrapping statistical method is used to compare the best performance improvement (lowest execution time) found by each initialisation technique after 100 generations. Listed parameters were chosen after initial exploratory experiments, although a full parameter sweep was beyond the scope of this investigation.

\begin{table*}
\centering
  \begin{tabular}{ll}
    Parameter & Value \\
    \hline
    Runs & 50 \\ 
    Population & 1000 \\
    Generations & 1000 for examples \\
      & 100 for initilisation comparison \\
    Initialisation & RHH, PI Grow \cite{fagan:2016:PI}, or Seed Only \\
    Crossover & Subtree \\
    Crossover Rate & 0.1 \\
    Mutation & Subtree \\
    Selection & Tournament \\
    Tournament Size & 2 \\
    Replacement & Generational \\
    Elitism & 100 Individuals \\
    Grammar & Perl Compatible Regex (PCRE) \\
  \end{tabular}
  \caption{GE Configuration}
  \label{tab:gpconfig}
\end{table*}

\vspace*{-9mm}

\subsection{Regex Problem Set}
\label{subsec:regex_test_suite}

We have assembled a benchmark suite of regex examples. These regexes can be considered ``Toy'' problems in that they are relatively short and understandable. The whole problem set can fit in this paper and does not pose a significant scaling challenge for GP. Results are not so complex as to be inscrutable. Experiments can be repeated and verified economically; a single run consisting of 1000 individuals across 1000 generations on a single core takes in the order of a couple of hours.

This benchmark set can however also be considered ``real-world'' as regexes are widely used in industry. Many programming languages have built-in support for regexes. 
Many of the chosen examples were taken from source code repositories of widely used libraries. Although larger amounts of code have been subjected to and improved with GP \cite{petke2014using,le2012systematic}, we see our benchmark suite as an introductory set of programs for experimenting with Genetic Improvement.

We gather regular expressions from online sources and popular Javascript libraries such as AngularJS \cite{AngularJS} and D3 \cite{D3}. We also take a problem from recent work \cite{bartoli2016inference} on evolving regular expressions in their entirety. Table~\ref{regexesuite} lists these regular expressions and we give a description and example input test data used for each problem.

\textbf{MAC address search}

\noindent The main use of regexes is for text extraction. We take an example which finds all MAC address instances in a string. This regex example was taken from the source code distributed with recent work in evolving this regex in its entirety \cite{bartoli2016inference,bartoliexample}. The 12 hexidecimal characters of a mac address can be upper or lower case and each pair of characters must be separated by a colon or minus sign. As mentioned in~\ref{subsec:regex_background}, an example test input string is:

\begin{center}
\texttt{eth0 119.63.193.196(5c:0a:5b:63:4a:82):4399}
\end{center}

\textbf{MAC address validation}

\noindent Regexes are frequently used to validate input data in client-side Javascript. We take one example here which validates a 12 character MAC address formatted as upper-case only, without any separator character. A string is valid if there are no characters before or after the MAC address.
An example test input value for this regex is:

\begin{center}
\texttt{5C0A5B634A82}
\end{center}

\textbf{Email validation}

\noindent Email addresses are frequently validated on websites before transmission server-side. This example is from a widely used Javascript library which supports validation of email addresses (AngularJS) \cite{AngularValidation}. No leading or trailing are characters are allowed.

\textbf{ISO 8601 datetime}

\noindent ISO 8601 is a standard for representing date and time strings and is also frequently validated. An example input string is:

\begin{center}
\texttt{2016-12-09T08:21:15.9+00:00}
\end{center}

\textbf{Scientific number}

\noindent From the same Javascript library as the preceding three, this validation regex checks for scientific number strings which can have leading plus or minus symbols and optionally exponent notation. Example input string is:

\begin{center}
\texttt{230.234E-10}
\end{center}

\noindent This regex is unusual for the purposes of validation as it allows leading and trailing whitespace. If this is not tested for in the test cases, then GP will remove this functionality. Similarly, edge cases such as a number less than one which does not contain a leading zero, e.g. ``\verb;.4536;'', must also be tested. 

\textbf{D3 interpolate number}

\noindent In another Javascript library \cite{D3} we found a regex similar to the previously mentioned Scientific number. This regex differs as its purpose is to extract matching strings of characters which are in scientific number format as opposed to validate them. The extracted number can be surrounded by other characters.  

\textbf{Catastrophic (QT3TS 1)}

\noindent
This problem was taken from a set of test cases designed to demonstrate interoperability between XML implementations. The regex was actually commented out of the test cases due to excessively high execution time. ``Catastrophic'' here refers to a regex which is obviously problematic in terms of run-time performance. Catastrophic Backtracking happens when part of regex matches text but a subsequent expression in the regex means the string is not a match. Regexes which are obviously problematic are clear targets for improvement. Where obvious improvements exist, problems like this one can be used as a test to validate the GP improvement system itself. A correct regex should extract ``\verb;bXcyXX;'' from the following string:

\begin{center}
\texttt{bbbbXcyXXaaa}
\end{center} 

\newpage
\textbf{Catastrophic (CSV P in 11th)}

\noindent This regex also exhibits Catastrophic Backtracking as it attempts to match only when the 11\textsuperscript{th} element of comma seperated string of values is an upper case P. The following input value should match:

\begin{center}
\texttt{1,2,3,4,5,6,7,8,9,10,11,P}
\end{center}

\textbf{Grammar parse rule}

\noindent Our final regex was taken from the PonyGE2 source code \cite{PonyGE2} and its purpose is to extract a production rule name and value from Backus–Naur form grammar. An example input string is:

\begin{center}
\texttt{<string> ::= <letter>|<letter><string>}
\end{center}

Note that this regex is also expected to extract values as groups with the names ``rulename'' and ``production''. Our fitness function does not currently test for capture groups. 

\begin{table}[h]
  \centering
  \caption{Test Suite of Regular Expressions (regex spanning more than one line should be concatenated)}
  \label{regexesuite}
  \setlength{\tabcolsep}{0.1em}
\begin{tabular*}{\textwidth}{lll}         
Name                                                          & Source\\  
  \hline                                                                                                                                                              
  MAC address search \cite{bartoli2016inference,bartoliexample} & \verb;([0-9A-Fa-f]{2}[:-]){5}([0-9A-Fa-f]{2}); \\  
  MAC address validation \cite{AngularValidation}               & \verb;^[0-9A-F]{12}$; \\
  Email validation \cite{AngularJS}                             & \verb;^(?=.{1,254}$)(?=.{1,64}@)[-!#$\%&'*+0-9=; \\ 
                                                                & \verb;?A-Z^_`a-z{|}~]+(\.[-!#$\%&'*+0-9=?A-Z^_`; \\ 
                                                                & \verb;a-z{|}~]+)*@[A-Za-z0-9]([A-Za-z0-9-]{0,61}; \\ 
                                                                & \verb;[A-Za-z0-9])?(\.[A-Za-z0-9]([A-Za-z0-9-]; \\
                                                                & \verb;{0,61}[A-Za-z0-9])?)*$; \\
  ISO 8601 datetime \cite{AngularJS}                            & \verb;^\d{4,}-[01]\d-[0-3]\dT[0-2]\d:[0-5]\d:; \\ 
                                                                & \verb;[0-5]\d\.\d+(?:[+-][0-2]\d:[0-5]\d|Z)$; \\  
  Scientific number \cite{AngularJS}                            & \verb;^\s*(-|\+)?(\d+|(\d*(\.\d*)))([eE][+-]?\d+); \\
                                                                & \verb;?\s*$; \\  
  D3 interpolate number \cite{D3}                               & \verb;[-+]?(?:\d+\.?\d*|\d*\.?\d+)(?:[eE][-+]?\d+)?; \\  
  Catastrophic (QT3TS 1) \cite{qt3ts}                           & \verb;.X(.+)+XX;  \\  
  Catastrophic (CSV P in 11th) \cite{csvcata}                   & \verb;^(.*?,){11}P;                                           \\  
  Grammar parse rule \cite{PonyGE2}                             & \verb;(?P<rulename><\S+>)\s*::=\s*(?P<production>; \\
                                                                & \verb;(?:(?=\#)\#[^\r\n]*|(?!<\S+>\s*::=).+?)+);   \\  
\end{tabular*}
\end{table}

\subsection{Input Test Data "Boosting"}
As mentioned in~\ref{subsec:fitness_function}, functionality error is a sum of match errors. To expand the set of input strings, we use a very basic approach to automatically generate additional test cases for each regex \cite{mcminn2004search,korel1990automated}. We take an input string and a known target regex which matches the desired text. We deterministically modify this input string until the target regex no longer matches. Any input strings which do not match are added to our test suite of input values. We modify input strings in two ways. We trim off characters from the start and end of the string until we find a string which does not match. We also replace every single character in the string with each character in the ranges 0-9 and a-Z. Replacing characters catches any special characters which are an important part of the problem. A variant evolved regex is only considered equivalent to the original seed regex in terms of this test data. 
Generally, the more terms there are in a regular expression, the more cases are needed to provide test coverage.

\section{Results}
\label{sec:results}

We initially report the results of a single GP run with a population of 1000 and 1000 generations. We use 1000 generations here to allow GP enough search time so that we may observe convergence in the population and hopefully find some interesting example ``best found'' regexes. 
Improvements were found in all regexes in our test suite as can be seen by the speedups listed in Table~\ref{table:results}.
We classify these improvements as ``specialisation'' as each regex is specific to the input test cases we used. The speedup values are indicative of typical improvements that were found for our chosen input data set. The speedup achievable also depends on the length and distribution of the input strings. In many cases, further testing is required to determine if the evolved regex is equivalent to the original seed used. The largest run-time improvements were found in the two examples of catastrophic backtracking with a particularly large performance improvement found for the ``Catastrophic (CSV P in 11th)'' problem.

A sample of improved regexes is shown in Table~\ref{table:exampleimprovements}. These examples appear complex as mutation has introduced what we consider ``noise'' - changes which have no measurable impact on runtime performance. From these results we can see that improvements are possible but that the result has become less readable. An additional ``minimisation'' GP run could be used to identify the minimal set of code changes which introduce a performance improvement enabling us to exclude superflous code changes \cite{weimer2009automatically}. Though a reasonable amount of each regex has been modified, the overall structure remains broadly the same. 

From these results, it would appear that the improvements require more than a single modification of a program. In other performance improvement work on much larger programs the number of changes to a program appears to result in less than a dozen changes amongst approximately 2000 lines of source code \cite{langdon:2015:optimizing}. In comparison, evolving a more terse language such as regular expressions appears to result in more of the original seed regex being modified.

\begin{table}[h]
  \centering
  \caption{``Best Found'' Regular Expression Speedups Found.}
  \label{table:results}
  \begin{tabular}{lr}
    Name                          & Speedup \\
    \hline
    MAC address search            & 1.14X   \\
    MAC address validation             & 1.57X    \\
    Email validation                   & 1.34X   \\
    ISO 8601 datetime               & 1.09X   \\
    Scientific number                  & 1.89X   \\
    D3 interpolate number              & 1.07X   \\
    Catastrophic (QT3TS 1)             & 3.60X   \\
    Catastrophic (CSV P in 11th)    & 112.64X \\
    Grammar parse rule                 & 1.31X   \\ 
  \end{tabular}
\end{table}

Of particulazr interest are the improvements at the extreme ends of what was found with GP.
What constitutes a valid MAC address is reasonably well and concisely specified as a sequence of exactly twelve uppercase letters (A-F) or numbers in the ``MAC address validation'' problem. It was not clear whether any improvement was possible for this problem and so we were surprised to see an improvement. Although the execution time was reduced, closer inspection reveals that the regex was specialised to the test cases used, as can be seen in Table~\ref{table:exampleimprovements}. This exemplifies the use of GP to explore the tradeoffs between functionality and performance. It is also likely that many of the longer more complex evolved regexes would not be considered semantically equivalent to the original regex.

\begin{table}[h]
  \caption{Example Regex Improvements.}
  \label{table:exampleimprovements}
  \setlength{\tabcolsep}{0.1em}
  \begin{tabular*}{\textwidth}{llr}
  Name                         & Example Improvement \\
  \hline
  MAC address search           & \verb;(..[:]){5,}[a-z\d(x{2277298,36899})+F]{3,653}; \\
  MAC address validation       &  \verb;^[0-9(?!9-G).]{12}$; \\
  Email validation             & \verb;^(?#.{1,254}$)(?=.{0,61}@)[-!#1A-Z*A-Z^_`a-z; \\ 
                               & \verb;{|?~]+(@.[M!#$%&'*Y0-9=?-Z^_`a-z{|~]+)i[A-Za-z; \\ 
                               & \verb;0-9]([A-Za-z0-9-]{0,61}[0-9a-za-z])?(\.[A-Za-z; \\ 
                               & \verb;0-9]([A-Za-z0-9-]{0,61}[A-Za-z0-9])?)*\w;\\
  ISO 8601 datetime            & \verb;^\d{4}-[01]\d-[0-3]\dT[0-2]\d:[0-5]\d:[0-5]\d\.; \\ 
                               & \verb;\d+([+][0-2]\d:[0-5]\d|^^A-Z[-5]d:[[]\d:[a-z]\d; \\ 
                               & \verb;:]{06227}.\d+2-b[%][58w07r\f-]u[{212,}]\?V0Z[1]; \\
                               & \verb;[!\w\d\d+][8dA]\F#2j[{2202,456}]\q|Z)$; \\
  Scientific number            & \verb;^\s*(-|\+)?(\d+|G\wp"\E,("s%Q+)?)([.Y]?\d+; \\
                               & \verb;([eE][+-]\d+)?\ *|)$; \\
  D3 interpolate number        & \verb;(?:\d+\.?\d*|[(-EQ)[]]*(?:\d+\.?[+E\/'3t\/++]?|; \\
                               & \verb;[[N0.]]*\F))(?:[eE][-+]?\d+)?; \\
  Catastrophic (QT3TS 1)       & \verb:.X([3-c{26,}]?)\wXX: \\ 
  Catastrophic (CSV P in 11th) & \verb:^([{}{6856945737,}5|{2555,}{,}WAT1][{137,2}v{3,}: \\
                               & \verb:v3|{1170,}|{20,}]*,){11}P: \\
  Grammar parse rule           & \verb;(?P<PPa><\S{6,6}>)\ ::=\s*(?P<OoFuDxisn>(tF6(); \\
                               & \verb;{,}A-Z(vPE)?|(?!<Plt>\s).{6}?){4}); \\ 
\end{tabular*}
\end{table}

In Table~\ref{table:catlineage} all improved regexes from a GP run are shown for the ``Catastrophic (QT3TS 1)'' problem. While the difference between the first two regexes clearly shows a 70\% reduction in execution time, the performance improvements thereafter are not as obvious. In regex number 3 the expression \verb;.+; (any character, any number of times) is replaced  by a set of characters listed between square brackets. The \verb;+; operator is greedy, allowing any number of characters to match so it is understandable that replacing this with something more specific may improve performance. The \verb;0-c; expression is evaluated as a range within square brackets. Replacing the match any (\verb;.;) operator with a range which more closely matches the input data example is also understandable. Despite the appearance of matching curly braces \verb;{76,}; these braces are not evaluated and are taken as individual characters within square brackets. The question mark allows zero or one of the characters in the set to be present.
We found a statistically significant difference between regex 2 and 3. 
Under further repeated tests comparing regex 3, 4 and 5 in Table~\ref{table:catlineage} no statistically significant difference was found between the running time of these regular expressions. This highlights a limitation with this approach whereby run-time performance variability can result in arbitrary mutations. 

In the ``Catastrophic (CSV P in 11th)'' problem a marked change can be seen after evolution. The original regex in Table~\ref{regexesuite} was 12 characters long. After evolution, it is 30 characters long in Table~\ref{table:exampleimprovements}. This variability is almost certainly a result of the difficulty of performing dependable benchmarking on software, particularly in interpreted environments.
Measuring instructions executed gives a deterministic dependable measure \cite{kuperberg2008bycounter} but counts each instruction the same. Instructions executed contain no information about how different combinations of instructions have varying execution time cost. In our GE system, each new individual regex is benchmarked for time. If some additional characters are added to a regex which do not measurably alter the running time of the regex, it is possible that the regex will receive a slightly lower running time than the original program. This demonstrates first-hand the difficulty in performing regex benchmarks. Although our inclusion of measuring runtime in the fitness function introduces a form of indirect parsimony pressure on the length of a regex, over many generations we still see neutral changes to the regex accumulate. 

\begin{table*}[h]
  \centering
  \caption{Catastrophic (QT3TS 1): Lineage of Improved versions.}
  \label{table:catlineage}
  \begin{tabular}{l|llr}
     & Regex & Time & Gen Found \\
    \hline
    1 & \verb;.X(.+)+XX;            & 0.00034746 & 0 \\
    2 & \verb;.X(.+)XX;             & 0.00010399 & 58 \\
    3 & \verb;.X([0-c{76,}]?)\wXX;  & 0.00009754 & 83 \\
    4 & \verb;.X([9-c{70,}]?)\wXX;  & 0.00009745 & 476 \\
    5 & \verb;.X([3-c{26,}]?)\wXX;  & 0.00009634 & 481 \\
  \end{tabular}
\end{table*}

\begin{figure*}
  \includegraphics[width=\textwidth]{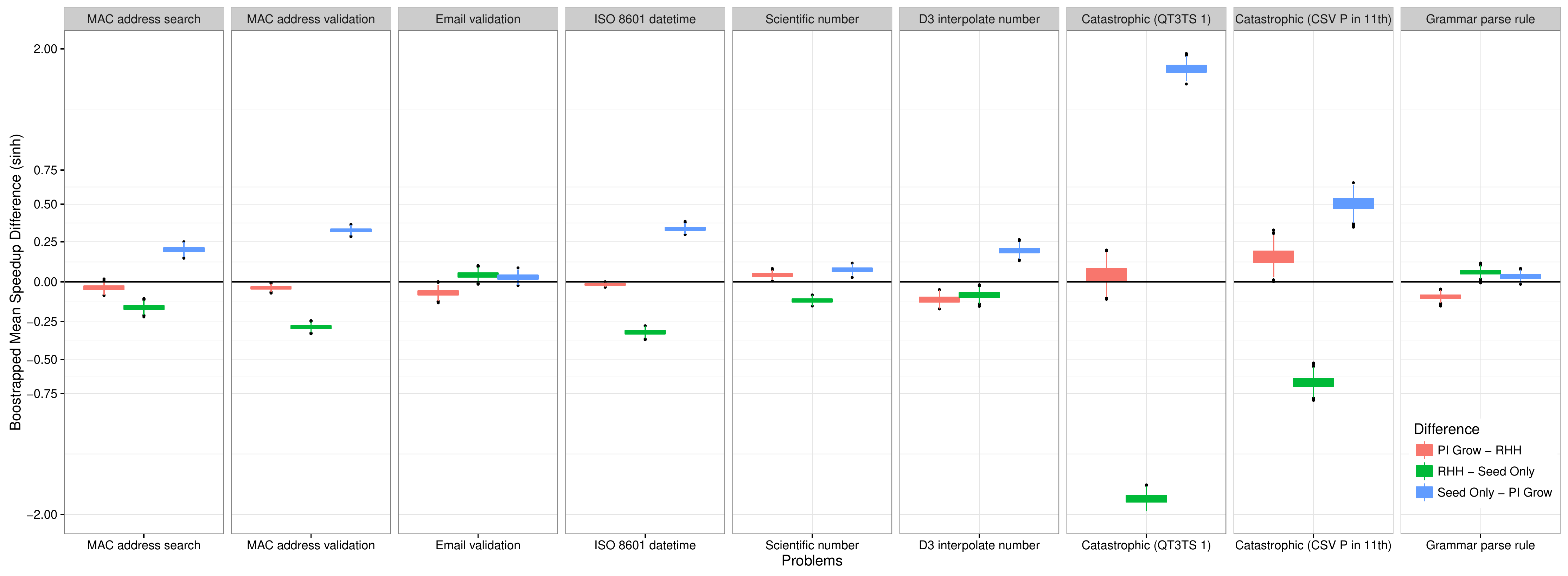}
  \caption{Comparison of Initialisation Methods.}
  \label{fig:initialisation_comparison}
\end{figure*}
  
In Figure~\ref{fig:initialisation_comparison} we compare the effect of the different initialisation methods as described in section~\ref{subsec:pop_seeding}. We perform 50 GP runs and use a bootstrapping method to gain statistical significance. We draw 1000 random samples (with replacement) from experiment values and find the difference between these values for each pairing of initialisation method. We take the mean of these 1000 differences. We repeat this process 1000 times to generate 1000 mean values of difference in speedup between each pairing. The difference between initialisation methods is statistically significant when the body of the boxplot does not cover the origin. \textbf{Seed Only} performed the best on 7 of the 9 problems with \textbf{RHH} performing best on the remaining two problems. This suggests that although we can increase diversity in the population, it does not always help in finding diverse implementations. \textbf{RHH} appears to outperform \textbf{PI grow} on 6 of the 9 problems, and their performance on the Catastrophic (QT3TS 1) problem is not clear. 

Though somewhat anecdotal, we did attempt evolution without the seed program. On one such run we were also able to find a regex which passed all our tests on the ``MAC address search'' problem without including the seed regex. This is worth investigating in future work as it appears that the evolution of regex from an initialisation mechanism such as ramped half-and-half or PI grow is within reach of a standard GP approach. Using standard GP appears to take a lot longer (hours instead of minutes) than multiple applications of a GP algorithm specialised for evolving regexes \cite{bartoli2016inference}.

\subsection{Limitations}

The performance improvements found are only valid within the input values used to test regex functionality correctness. Where some functionality provided by the original regex is not specifically tested by the input strings we can assume GP will likely remove such functionality. If removing this functionality does not cause a fitness penalty then it is likely to be removed if it reduces the execution time of the regex. Test suite coverage as well as determining program equivalence is a recognised problem \cite{barr2015oracle} and although we find performance improvement, we can only claim the improvements as specialisations of the regexes to the input data. In other words, we maintain functionality only as far as is measured by our test strings. If the tests don't measure it, functionality is removed during experiments.
There remains the possibility that an evolved regex which is correct for the input test data is not fully equivalent to the original regex\footnote{There is a larger open question here on how to fully test program equivalence between original seed and evolved regex. In practice, if GP can be used to find a performance improvement, we may ask a human programmer to decide whether the evolved regex can be used as a replacement.}.  
Though our benchmark suite is compact, the power of our results to recommend one initialisation method over another may only be used as a guide for future work due to the relatively small size of the benchmark suite.

\section{Conclusions}
\label{sec:conclusions}

We have presented and experimented with a compact benchmark test suite of ``real world'' regex examples.
Our evaluation demonstrates the utility of Grammatical Genetic Programming in exploring the trade-off between functionality and performance. We were able to find improvements in code from widely used systems. This answers our first research question from \ref{sec:intro} as we readily find regexes ``in-the-wild'' for which we can find some performance improvement. As to our second research question as to whether or not we can maintain functionality, we are somewhat undecided. For the shortest regexes, we were able to find improvements, but these improvements clearly did not maintain equivalent functionality. For the more complex regexes which we evolved, we found improvements but cannot say for sure if these improvements deleteriously affected functionality. 

We were also able to use our GE system to find regex improvements in the GE system itself, demonstrating an initial step towards a self-improving GP system \cite{cody2015locogp}.  
From our results it appears that it is not a problem to use only a seed program as the initial starting point in evolving improvement. Improvements found are incremental in that the overall structure of the original regex is maintained. As such, other methods for finding increasingly different regexes should be explored.

\subsection{Future Work}
\label{subsec:future_work}

A common thread in this future work is the notion of building up a library or repository of known, well understood regexes. 
Regex snippets could be continually gathered as they are parsed \cite{ReverseGE} from human-written examples as well as over the course of evolutionary runs. Within a GP context, this library could be used to drive a semantic crossover operator.

An approach to elicit interesting regex variants is to use many evolutionary runs with different GP configurations \cite{feldt1998generating}. We envision a hierarchical approach whereby progressively larger numbers of regex primitives are excluded from each GP run. Excluding a primitive from a regex and the grammar would force GP to find a functionally equivalent regex without using that particular primitive. Repeating this for all individual primitives in the known regex should yield a set of alternative regex implementations, which can also be added to our library. The process can be repeated for all pairs of primitives, and so on for triplets etc. 

To gain a more representative analysis of real-world regexes, the benchmark suite can be expanded to include thousands of regexes \cite{chapman2016exploring}. As effort has been made to discover the most prevalent and effective performance improvements in Javascript \cite{selakovic2016performance}, a similar approach may be useful across a large corpus of regexes. If common regex patterns are found we may apply regex improvements more deterministically. If we can gain a broad understanding of what functionality is most often required, and also elicit the most common improvement opportunities in human-written regexes, we may be able to define a canonical set of the ``best'' regexes to use in the majority of cases, thus providing more ``bang-for-buck'' to the programmer \cite{gallagher2008program}. The goal is to reduce time spent writing and debugging regexes for use cases which are well understood and, in some sense, already ``solved''.

\section*{Acknowledgments}

This research is based upon works supported by Science Foundation Ireland under grant 13/IA/1850 and 13/RC/2094 which is co-funded under the European Regional Development Fund through the Southern \& Eastern Regional Operational Programme to Lero - the Irish Software Research Centre (\url{www.lero.ie}).

\bibliographystyle{acm}
\bibliography{sigproc} 

\end{document}